\documentclass[runningheads]{llncs}
\usepackage{graphicx}
\usepackage{amssymb}
\usepackage{booktabs}
\usepackage{pbox}
\usepackage{bm}
\usepackage{subcaption}
\usepackage{paralist, tabularx}
\usepackage{multirow}
\usepackage{array}
\usepackage{bbm}
\usepackage{float}
\usepackage{comment}
\usepackage{pifont}
\usepackage{color}
\usepackage{hyperref}
\usepackage{marvosym}
\usepackage{ifsym}
\usepackage{orcidlink}
\newcommand{\cmark}{\ding{51}}
\newcommand{\xmark}{\ding{55}}

\begin{document}
\title{Learning Efficient Representations for Image-Based Patent Retrieval}
%
\author{Hongsong Wang\inst{1}\textsuperscript{*}\textsuperscript{(\Letter)}\orcidlink{0000-0002-9464-1778} \and
	Yuqi Zhang\inst{2}\textsuperscript{*}\orcidlink{0000-0001-7094-3838}}
\authorrunning{H. Wang and Y. Zhang}
%
\institute{Department of Computer Science and Engineering, \\Southeast University, Nanjing 210096, China \and
	Baidu Inc \\
	\email{hongsongwang@seu.edu.cn, zhangyuqi1991@gmail.com}}

\maketitle              
\begin{abstract}
Patent retrieval has been attracting tremendous interest from researchers in intellectual property and information retrieval communities in the past decades. However, most existing approaches rely on textual and metadata information of the patent, and content-based image-based patent retrieval is rarely investigated. Based on traits of patent drawing images, we present a simple and lightweight model for this task. Without bells and whistles, this approach significantly outperforms other counterparts on a large-scale benchmark and noticeably improves the state-of-the-art by 33.5\% with the mean average precision (mAP) score. Further experiments reveal that this model can be elaborately scaled up to achieve a surprisingly high mAP of 93.5\%. Our method ranks first in the ECCV 2022 Patent Diagram Image Retrieval Challenge. 

\end{abstract}
\footnotetext{\textsuperscript{*}{Equal Contribution. \textsuperscript{\Letter}Corresponding author.}}

\section{Introduction}
\label{sec:intro}
With a large and ever-increasing number of scientific and technical documents on the web every year, diagram image retrieval and analysis~\cite{yang2020diagram,bhattarai2020diagram} become crucially important for intelligent document processing and understanding. Different from natural images that contain rich information about color, texture, and intensity, diagram images only involve line and shape. Although existing computer vision methods have achieved significant progress on natural image understanding (e.g., classification, detection and retrieval)~\cite{he2016deep,tan2019efficientnet,liu2023efficient,wang2021afan}, diagram image understanding retains a challenging and less developed area. 

\begin{figure}[t]
	\centering
	\begin{subfigure}[b]{0.55\linewidth}
		\includegraphics[width=\textwidth]{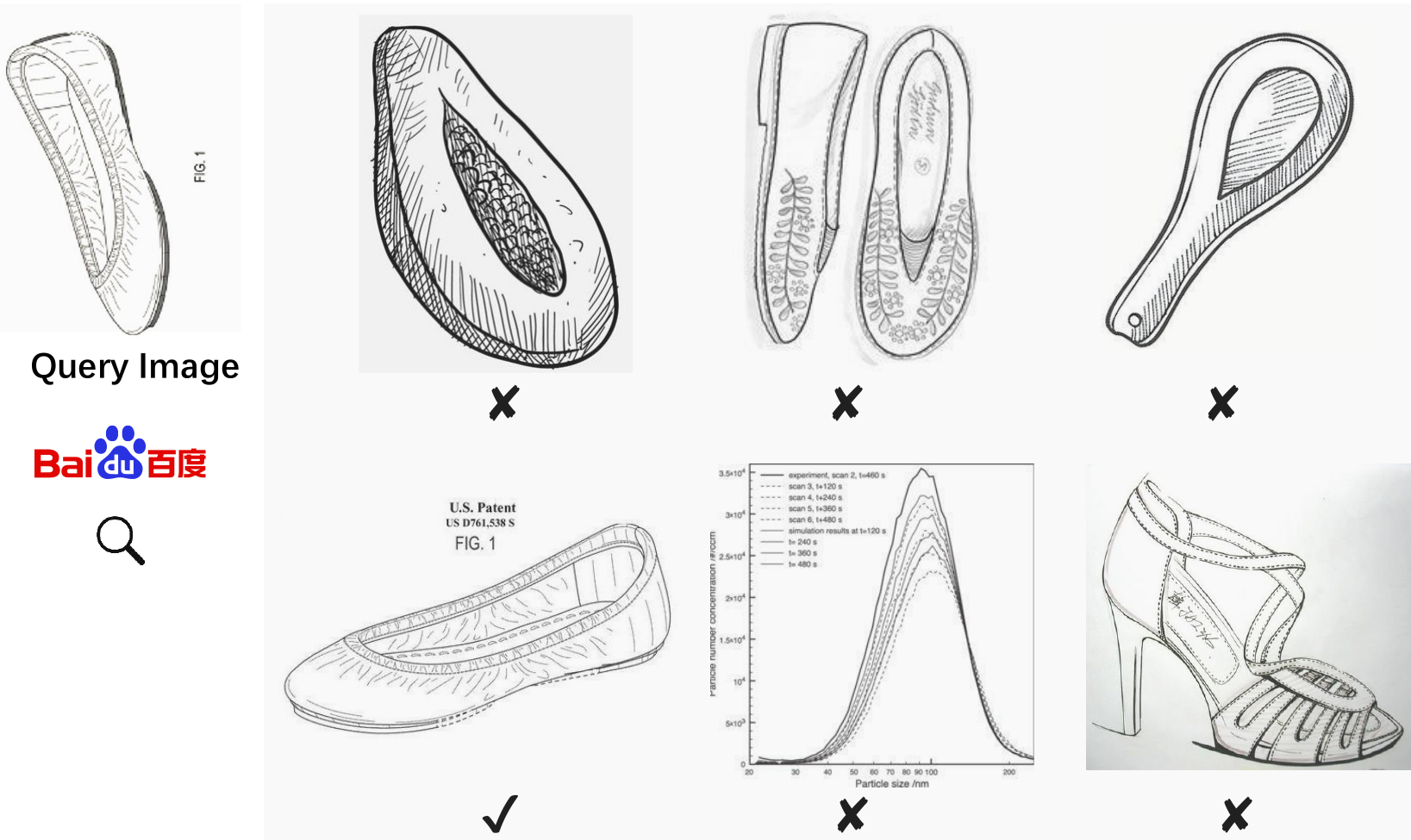}
		\caption{Image-based patent retrieval}
	\end{subfigure}
	\begin{subfigure}[b]{0.44\linewidth}
		\includegraphics[width=\textwidth]{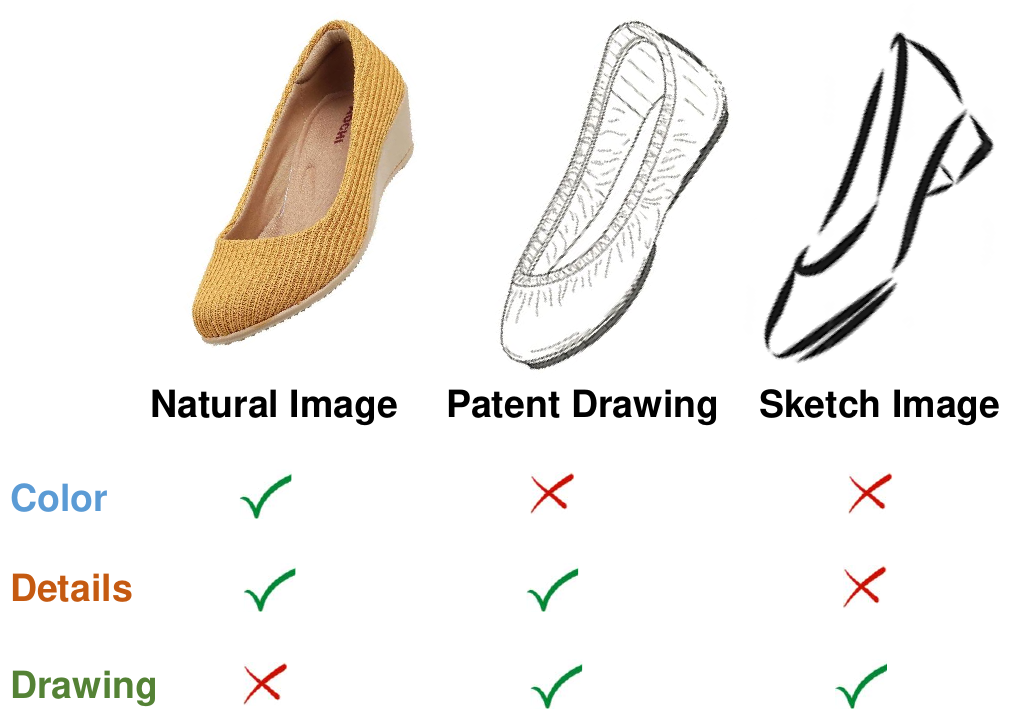}
		\caption{Different types of images}
	\end{subfigure}
	\caption{An illustration of image-based patent retrieval.}
	\label{fig:patent}
\end{figure}

Image-based patent retrieval~\cite{hanbury2011patent,bhatti2013image,zhang2023graph}, one of the most typical research topics of diagram image understanding, has drawn increasing interest from intellectual property and information retrieval communities. The need for effective and efficient systems of patent retrieval becomes inevitably crucial due to the tremendous amounts of patent data. As visual image plays an essential role in information retrieval, patent image search can help people quickly understand the patent contents and check differences between patents. However, accurate patent retrieval based on visual content remains an open challenge. Commercial search engines such as Google and Baidu currently fail to retrieve relevant patent images with the query of a patent drawing~\cite{kucer2022deeppatent}. An example of this task is illustrated in Fig.~\ref{fig:patent}(a), where many unrelated drawing images with a similar style to the query are retrieved. 

One of the well-concerned problems in computer vision most related to image-based patent retrieval is sketch-based image retrieval~\cite{eitz2010sketch}. 
Although both are abstract drawings, there are subtle differences between sketch images and patent drawings. As shown in Fig.~\ref{fig:patent}(b), patent drawings contain more detailed contours and patterns about the object when compared with sketch images. 

Existing patent search systems mostly use textual and metadata information. 
Only little published work studies content- and image-based patent retrieval. These approaches represent the patent drawing with relational skeletons~\cite{huet2001relational}, edge orientation autocorrelogram~\cite{tiwari2004patseek}, contour description matrix~\cite{zhiyuan2007outward}, adaptive hierarchical density histogram~\cite{vrochidis2010towards}, etc. 
For example, the patent retrieval system PATSEEK~\cite{tiwari2004patseek} applies shape-based retrieval and represents a patent image with an edge orientation autocorrelogram.
However, these representations are low-level and handcrafted visual features, which are not discriminative enough for large-scale applications. In addition, previous public patent retrieval datasets such as the \texttt{concept}~\cite{vrochidis2012concept} only contains a limited number of ID categories and images. The evaluation protocols of image-based patent retrieval are also not consistent across most existing approaches.

Recently, a large-scale dataset called \texttt{DeepPatent}~\cite{kucer2022deeppatent} focused on patent drawings is introduced. This dataset makes it possible for large-scale deep learning-based image-based patent retrieval. The PatentNet~\cite{kucer2022deeppatent} is currently the only deep learning model designed for large-scale image-based patent retrieval. However, this model is sophisticatedly trained in two stages with different training losses. To the best of our knowledge, there exists no single-stage training model that exploits the unique characteristics of patent drawings and comprehensively studies how to train an effective patent retrieval system from patent images.

To this end, we present a simple yet strong pipeline for image-based patent retrieval which adopts a lightweight backbone and subsequently applies a neck structure to obtain low-dimensional representation for efficient patent retrieval. Training this network is straightforward by only using a classification loss in a geometric angular space and data augmentation without scaling, which is specifically designed for patent drawings according to their characteristics. Experimental results demonstrate that the proposed method significantly outperforms both the traditional and deep learning-based counterparts. In addition, this baseline can be easily scaled up from the current lightweight model to large models which achieve even better performance. 

\section{Related Work} \label{sec:related_work}
\subsection{Image-Based Patent Retrieval}
Patent retrieval~\cite{krestel2021survey} is the task of finding relevant patents for a given image query, which can be a sketch, a photo or a patent image. An image and text analysis approach automatically extracts concept information describing the patent image content~\cite{vrochidis2012concept}. A retrieval system uses a hybrid approach that combines feature extraction, indexing and similarity matching to retrieve patent images based on text and shape features~\cite{mogharrebi2013retrieval}. A combined method of classification and indexing techniques retrieves patents for photo queries and evaluates it on a large-scale dataset~\cite{gong2020image}.
PatentNet~\cite{kucer2022deeppatent} learns to embed patent drawings into a common feature space using a contrastive loss function, where similar drawings are close and dissimilar drawings are far apart. 
The diversity and complexity of patent images which contain different objects and the data scarcity and imbalance are main difficulties of image-based patent retrieval. 

\subsection{Sketch-Based Image Retrieval}
Sketch-based image retrieval~\cite{xu2022deep} receives hand-drawn sketch as an input and retrieves images relevant to the sketch. 
Existing deep learning approaches can be roughly divided into three categories: generative models, cross-modal learning and zero-shot learning. Generative models aim to generate synthetic images or sketches from the input sketch or image using generative neural networks~\cite{chen2018sketchygan,bhunia2021more}. Cross-modal learning approaches learn a common feature space for sketches and images using techniques such as metric learning, domain adaptation and knowledge distillation~\cite{chaudhuri2020crossatnet,sain2021stylemeup}. Zero-shot learning retrieves images that belong to unseen categories using semantic information such as attributes or word embeddings~\cite{yelamarthi2018zero,verma2019generative}. This problem is extremely challenging as the appearance gap between sketches and natural images makes it difficult to learn effective cross-modal representations and style variations among different sketchers introduce noise and ambiguity to the sketch representations. In contrast, image-based patent retrieval does not involve cross-modal learning and variations among patent drawings are smaller than those among sketches.

\section{Method}
\begin{figure*}[t]
	\centerline{
		\includegraphics[width=0.95\linewidth]{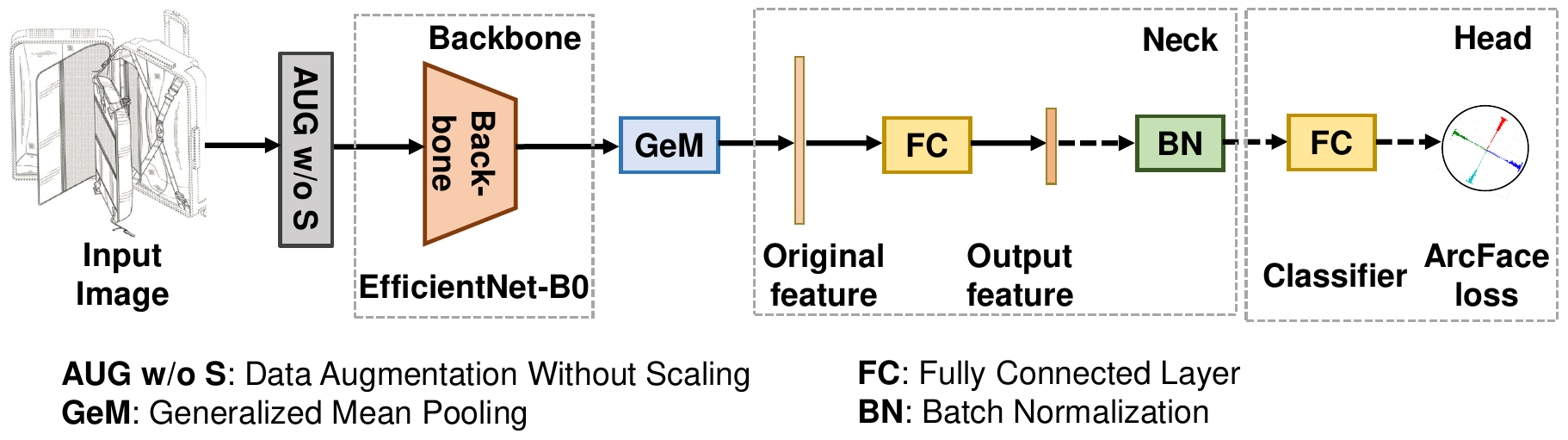}}
	\caption{The network structure of the proposed method. Dotted lines denote modules that only appear in the training pipeline.} 
	\label{fig:method}
	\vspace{-0.1in}
\end{figure*}

We present a simple yet effective training pipeline for image-based patent retrieval, which is shown in Fig.~\ref{fig:method}. Similar to the strong baseline for person re-identification (Re-ID)~\cite{luo2019strong}, the network mainly consists of the backbone, the neck and the head. However, training strategies and structures of key components are very different from those of Re-ID. Details are described as follows. 

\label{sec:method}

\subsection{Backbone Network}
EfficientNet-B0~\cite{tan2019efficientnet} is an effective and efficient convolutional network obtained by neural architecture search~\cite{zoph2017neural,tan2019mnasnet}.
Compared with ResNet-50~\cite{he2016deep}, EfficientNet-B0 significantly reduces network parameters and FLOPS by an order of magnitude. The main building block of this network is the mobile inverted bottleneck called MBConv~\cite{sandler2018mobilenetv2,tan2019mnasnet}, which uses the efficiency inverted structure in the residual block. This architecture can be easily scaled up from dimensions of depth, width, and resolution to obtain a family of models.

Given the output feature maps of the backbone network, we use Generalized Mean (GeM) pooling~\cite{radenovic2018fine} instead of the commonly used max- or average-pooling to get image features. The pooled feature dimension is the same as the number of channels of feature maps. Let $f \in \mathbb{R}^{n}$ be the output feature vector of dimension $n$. For the $k$-th component of vector $f$, the formulation of GeM pooling is 
\begin{equation}
	{f_k} = {(\frac{1}{{{\rm{|}}{{\rm Z}_{\rm{k}}}{\rm{|}}}}{\sum\limits_{x \in {{\rm Z}_{\rm{k}}}} x ^{{p_k}}})^{{1 \mathord{\left/
					{\vphantom {1 {{p_k}}}} \right.
					\kern-\nulldelimiterspace} {{p_k}}}}},
\end{equation}
where ${\rm Z}_{\rm{k}}$ is the $k$-th feature map, and $p_k$ is a learnable pooling parameter, $p_k > 0$. This operation is average-pooling when $p_k=1$, and approximates max-pooling when $p_k \to \infty$. 
The GeM pooling focuses on the salient and discriminative features from different spatial positions of the image.


\subsection{Neck Structure}
The strong baseline for Re-ID presents a neck structure that consists of a batch normalization (BN) layer and empirically analyzes the functions of this BN layer while combining ID classification loss and triplet loss~\cite{luo2019strong}. There exists inconsistency between the classification loss and the triplet loss while combing them in the same feature space, and the BN layer alleviates this inconsistency by separating the two losses into different feature spaces.

Different from previous neck structures~\cite{luo2019strong,ye2021deep} that involve solely a BN layer for person re-identification, we design a neck structure that comprises a fully connected (FC) layer and a consecutive BN layer. The FC layer maps the pooled feature vector of an image into a compact and low-dimensional space, and, as a result, reduces the parameters of the classifier. In addition, it transforms feature vectors of various dimensions from different backbones to fixed dimensions. 

During training, the BN layer normalizes the feature distributions and improves the intra-class compactness and inter-class discrimination, which benefits the ID classification loss. In the inference phase, the output feature of the FC layer is directly used for image retrieval.

\subsection{Classification Head}
The PatentNet~\cite{kucer2022deeppatent} first trains the patent retrieval model with the classification loss and then finetunes the network using either the triplet or contrastive loss. The strong baseline~\cite{luo2019strong} trains the person re-identification network with both the classification and triplet losses. Our approach for image-based patent retrieval is trained in a single stage and only uses classification loss.

Let $x_i$ denote the retrieval feature vector of the $i$-th image sample, and $y_i$ be the patent ID label. Suppose there are $m$ different ID categories and $N$ samples, the widely used cross-entropy loss or the Softmax loss $L_{\rm{c}}$ is 
\begin{equation}
	{L_{\rm{c}}} =  - \frac{1}{N}\sum\limits_{i = 1}^N {\log \frac{{{e^{W_{{y_i}}^T{x_i} + {b_{{y_i}}}}}}}{{\sum\nolimits_{j = 1}^m {{e^{W_j^T{x_i} + {b_j}}}} }}},
\end{equation}
where $W$ and $b$ are the weight and bias parameters of the classifier that consists of an FC layer.

The ArcFace loss~\cite{deng2019arcface} first performs L2 normalization upon the feature vector and the weight of the classifier, then computes the cosine distance in a geometric angular space. The loss $L_a$ is formulated as
\begin{equation}
	{L_a} =  - \frac{1}{N}\sum\limits_{i = 1}^N {\log \frac{{{e^{s \cdot \cos ({\theta _{{y_i}}} + m)}}}}{{{e^{s \cdot \cos ({\theta _{{y_i}}} + m)}} + \sum\nolimits_{j = 1,j \ne {y_i}}^m {{e^{s \cdot \cos {\theta _j}}}} }}},
\end{equation}
where $s$ and $m$ are scale and angular margin penalty hyperparameters, respectively. 
As discussed in~\cite{deng2019arcface}, the ArcFace loss could enhance similarity for intra-class samples and increase diversity for inter-class samples, which is particularly suitable for large-scale retrieval scenarios due to the huge intra-class appearance variations. 

\begin{figure}[t]
	\centerline{
		\includegraphics[width=0.95\linewidth]{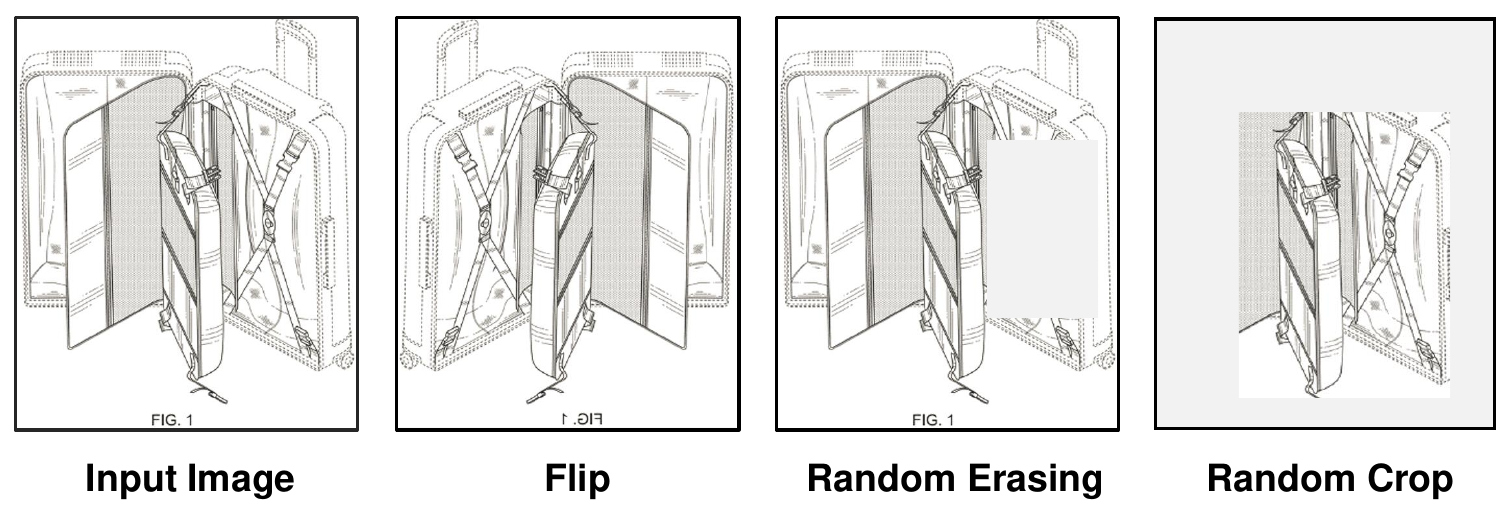}}
	\caption{Typical data augmentations without scaling transformation for image-based patent retrieval. For \texttt{Random Crop}, the cropped image is not resized to the original image resolution.} 
	\label{fig:augmentation}
	\vspace{-0.1in}
\end{figure}
\subsection{Data Augmentation Without Scaling} \label{subsec:aug}
Scaling Transformation is a prevalent data augmentation technique for natural image recognition and analysis. As the concerned object in images varies considerably in different sizes, scaling transformation makes the model learn scale-invariant features from images. For image understanding with deep learning, scaling transformation is almost inevitably included in any data augmentations.

Different from natural images, patent drawings span a relatively fixed line width in the document, regardless of the content, and so objects in the drawings vary little in size. For this reason, we present data augmentation without scaling (\texttt{AUG w/o S}) for image-based patent retrieval. Typical \texttt{AUG w/o S} transformations are shown in Fig.~\ref{fig:augmentation}. Empirically studies of data augmentations are provided in experiments.

\section{Experiments}

\subsection{Datasets and Implementation Details}
The DeepPatent~\cite{kucer2022deeppatent} is a recent large-scale dataset for image-based patent retrieval. It consists of a total of 45,000 different design patents and 350,000 drawing images. The object in a patent is presented in multiple drawing images from different views. The training set has 254,787 images from 33,364 patents, and the validation set has 44,815 images from 5,888 patents. This dataset is dedicated to visual patent retrieval based on detailed and abstract drawings. 
The mean Average Precision (mAP) and ranked accuracy (Rank-N) are used to evaluate the performance of retrieval methods. 

The input image resolution is 256$\times$256 if not otherwise specified. The output feature dimension of the FC layer in the neck structure is 512. The scale and margin hyperparameters $s$ and $m$ in the ArcFace loss are 20 and 0.5, respectively. 
The network is trained using the AdamW optimizer with an initial learning rate of 0.001 and weight decay of 0.0005. The batch size is 128, and the maximum number of iterations is 20,000.

\subsection{Experimental Results}
\begin{table}[t!]
	\centering
	\caption{Comparison with the state-of-the-art approaches for large-scale patent retrieval on the DeepPatent dataset.}
		\begin{tabular}{l|cccc}
			\toprule
			Method & mAP  & Rank-1  & Rank-5 & Rank-20 \\
			\midrule
			HOG~\cite{dalal2005histograms} & 8.3  &  27.2  &  31.7  & 35.9 \\
			SIFT FV~\cite{csurka2011xrce} & 9.2  & 20.6  & 28.9   &  37.5 \\
			LBP~\cite{ojala2002multiresolution} & 6.9 & 21.0  & 25.2  &  34.3 \\
			AHDH~\cite{vrochidis2010towards} & 9.5 & 28.8  & 34.3 & 39.9 \\
			VisHash~\cite{oyen2021vishash} & 9.3 & 27.4  & 34.0  & 40.2 \\
			\midrule
			Sketch-a-Net~\cite{yu2017sketch} & 13.5 & 36.1  & 45.1  & 53.6 \\
			PatentNet~\cite{kucer2022deeppatent} & 37.6  & 69.1 & 78.4  & 84.1 \\
			\midrule
			Ours  & 71.2  & 88.9  & 95.8 & 98.1 \\
			Ours w/o ArcFace & 55.4 & 82.3 & 91.3  & 95.2 \\
			\bottomrule
		\end{tabular}
	\label{tab:patent_results}
\end{table}

As image-based patent retrieval is a relatively new research topic and traditional general techniques of computer vision are likely to perform well, we compare our approach with both deep learning and traditional approaches.

Representative traditional methods are HOG~\cite{dalal2005histograms}, SIFT FV~\cite{csurka2011xrce,perronnin2010improving}, AHDH~\cite{vrochidis2010towards}, and VisHash~\cite{oyen2021vishash}.
Deep learning-based comparison methods are Sketch-a-Net~\cite{yu2017sketch} and PatentNet~\cite{kucer2022deeppatent}. 
The proposed method significantly outperforms both traditional and deep learning approaches. 
It beats the recently proposed deep learning-based PatentNet~\cite{kucer2022deeppatent} by 33.5\% and 19.8\% for the scores of mAP and Rank-1, respectively. 

\subsection{Analysis and Visualization}
\begin{figure*}[t]
	\centering
	\begin{subfigure}[b]{0.42\linewidth}
		\includegraphics[width=\textwidth]{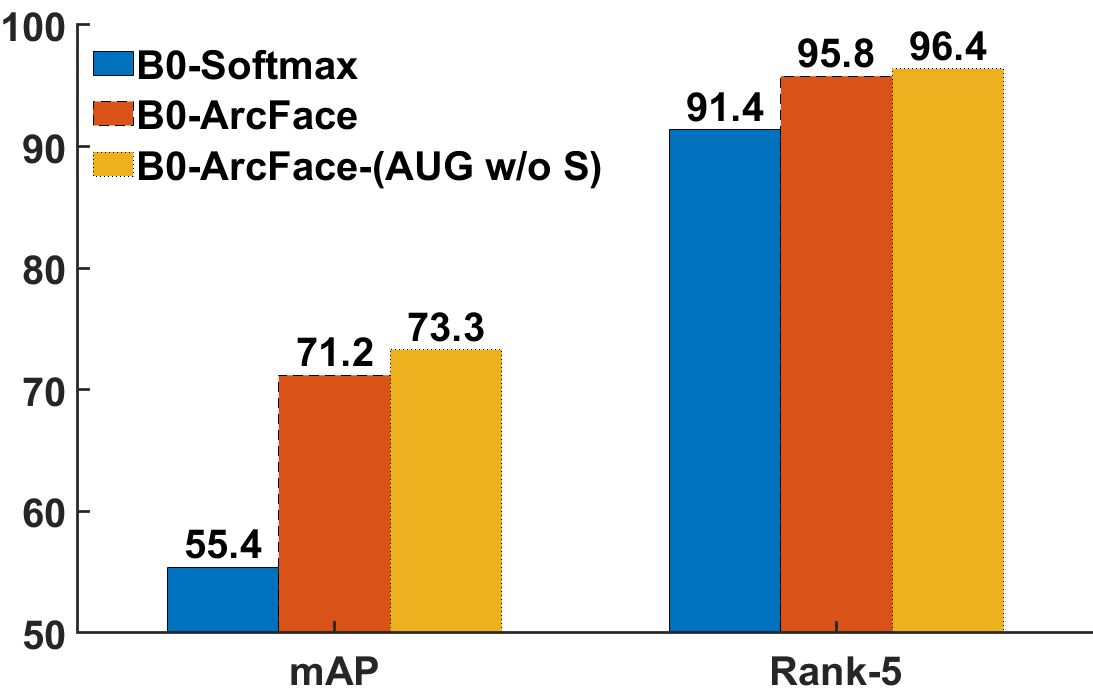}
		\caption{Ablation studies}
	\end{subfigure}
	\hspace{0.1in}
	\begin{subfigure}[b]{0.42\linewidth}
		\includegraphics[width=\textwidth]{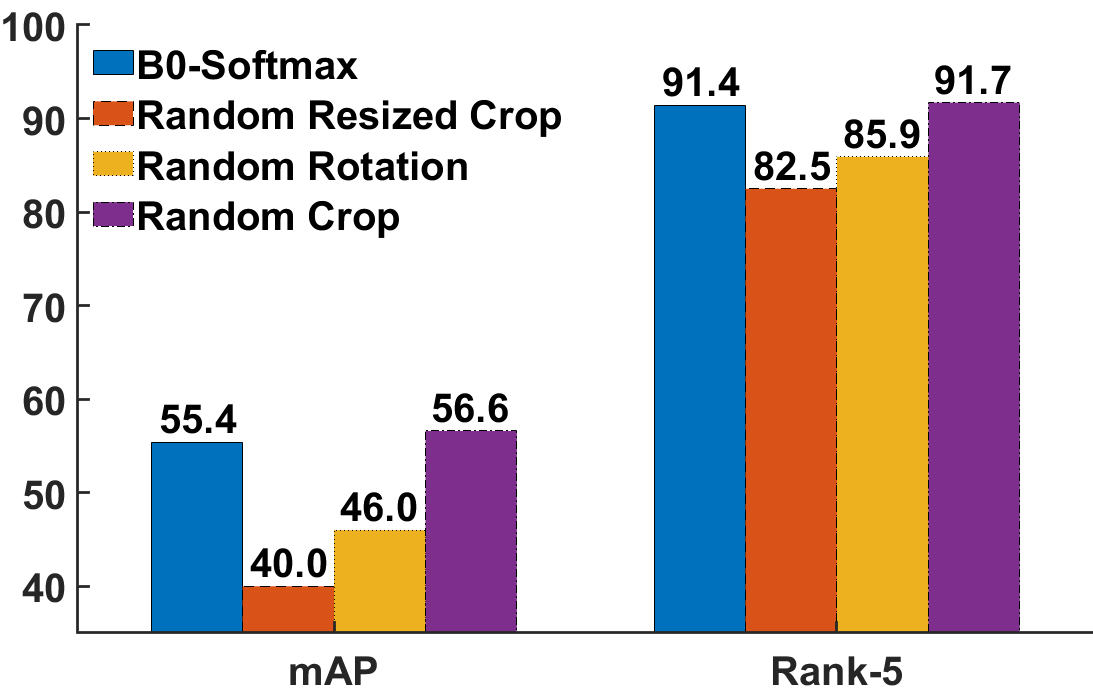}
		\caption{Effects of augmentations}
	\end{subfigure}
	\caption{Effects of key components of the proposed method.}
	\label{fig:ablations}
\end{figure*}
\begin{figure*}[t]
	\centering
	\begin{subfigure}[b]{0.42\linewidth}
		\includegraphics[width=\textwidth]{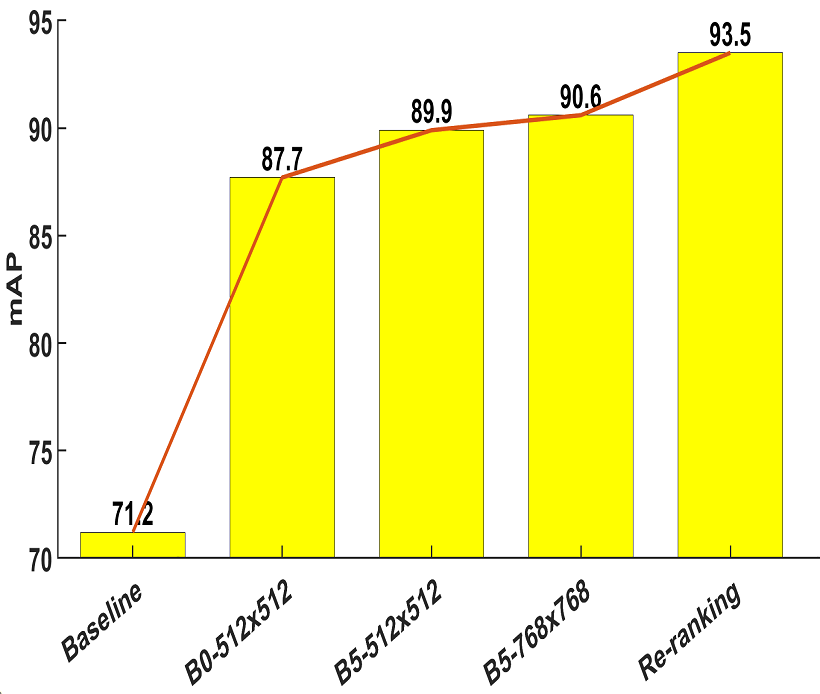}
		\caption{Ablation studies}
	\end{subfigure}
	\hspace{0.1in}
	\begin{subfigure}[b]{0.42\linewidth}
		\includegraphics[width=\textwidth]{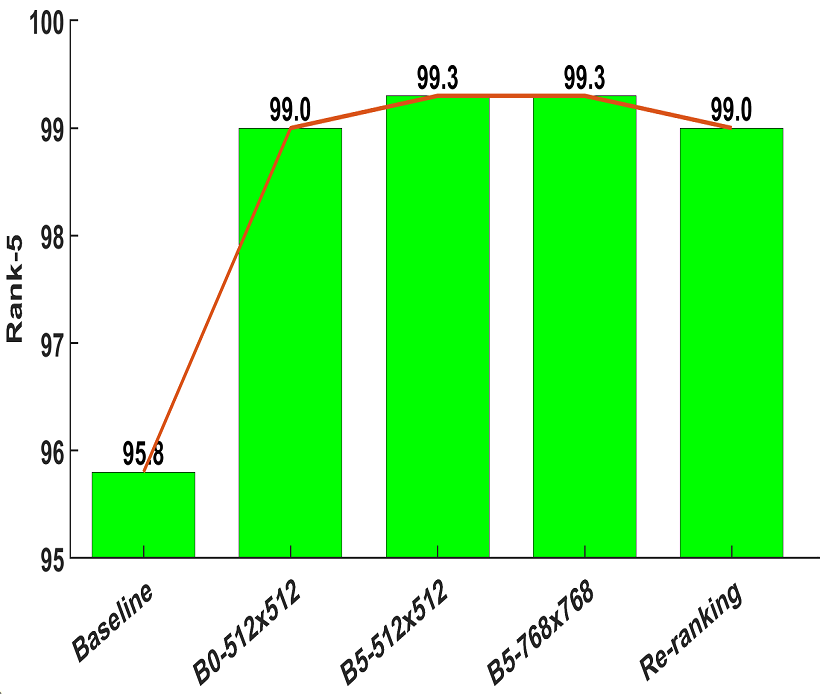}
		\caption{Effects of augmentations}
	\end{subfigure}
	\caption{Scalability analysis of the proposed method. For the sake of simplicity, \texttt{B0} and \texttt{B5} denote EfficientNet-B0 and EfficientNet-B5, respectively, and $m \times m$ means that the input resolution is $m \times m$.}
	\label{fig:scalability}
\end{figure*}

Ablation studies of key components of the proposed method are shown in Fig.~\ref{fig:ablations}(a). ArcFace loss significantly improves the results over Softmax loss for large-scale image-based patent retrieval, as it increases the value of mAP from 55.4\% to 71.2\%. Data augmentation without scaling (\texttt{AUG w/o S}) further boosts the performance over the proposed strong baseline \texttt{B0-ArcFace} which adopts EfficientNet-B0 as the backbone and uses the ArcFace classification loss.

In Fig.~\ref{fig:method}, \texttt{AUG w/o S} are specifically designed in the proposed method. Here we experimentally evaluate different augmentations techniques and provide results in Fig.~\ref{fig:ablations}(b). Surprisingly, we find that the commonly used \texttt{Random Resized Crop} severely hurts the performance while \texttt{Random Crop} without scaling could improve performance. The results confirm our hypothesis in Subsection~\ref{subsec:aug} that scaling transformation is not suitable for image-based patent retrieval. Another observation is that \texttt{Random Rotation} also dramatically damages the performance. One possible reason is that drawing images are horizontally placed in the document without any geometric distortion.

Although state-of-the-art performance has been reached, the proposed strong baseline can be scaled up to achieve even stronger performance. As a trial experiment, we extend the model from three aspects: increasing input resolution, using a large backbone, and post-processing with re-ranking. The results are summarized in Fig.~\ref{fig:scalability}. 
Notably, the mAP score is first boosted from 71.2\% of the baseline to 87.7\% with the input image resolution of $512 \times 512$, then gets 89.9\% using EfficientNet-B5, and finally reaches 93.5\% by combining these three expanding techniques.
Notably, the mAP score is astonishingly boosted from 71.2\% of the baseline to 93.5\% by combining these three expanding techniques. 
The final extended approach wins the first place in the Image Retrieval Challenge of an International Conference. 

\begin{figure*}[t]
	\centering
	\includegraphics[width=\textwidth]{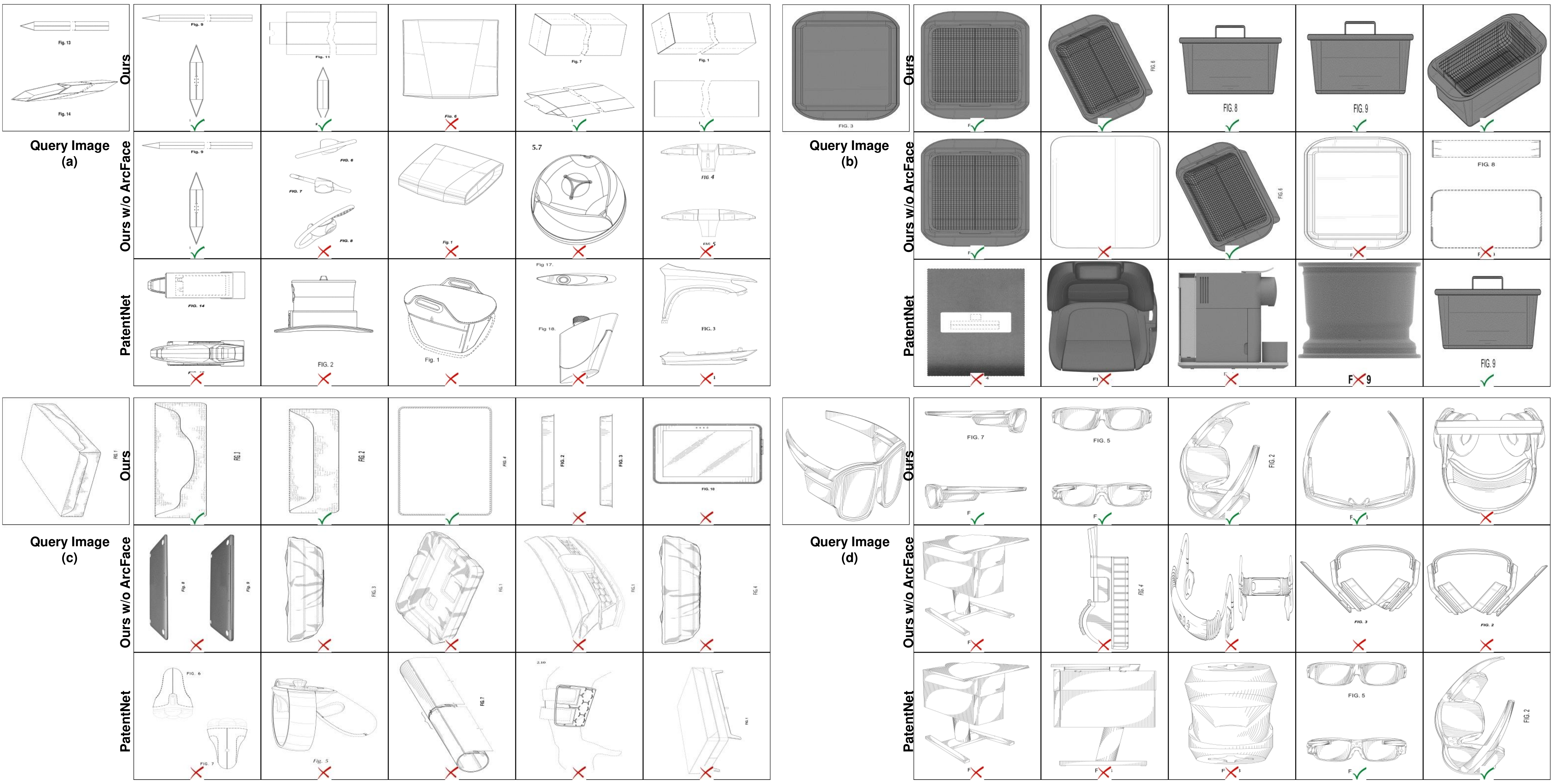}
	\caption{Comparison of image-based patent retrieval results of different approaches of ours and the state-of-the-art PatentNet~\cite{kucer2022deeppatent}. Given a query image, we only show the top-5 retrieved images. Symbols~\cmark~and~\xmark~denote correct and wrong hits, respectively.}
	\label{fig:visulize}
	\vspace{-0.1in}
\end{figure*}

To further investigate the retrieval performance, we visualize the search results of different approaches in Fig.~\ref{fig:visulize}. We find that our approach is robust to viewpoints. For the query patent (b) which is an aerial view of a patent object, our approach successfully retrieves the patent images from other different views while the recent state-of-the-art PatentNet~\cite{kucer2022deeppatent} fails to find this patent for the top-4 retrieved results. For the query patent (c) which contains apparent texture patterns, the PatentNet~\cite{kucer2022deeppatent} find other patents with similar shapes by mistake while ours correctly retrieves the right patents from other views. 

\section{Conclusion}
In this paper, we empirically study image-based patent retrieval and present a simple, lightweight yet strong model which consists of a backbone network, neck structure, and training head. 
We find that the loss function plays an important role in this task, and deliberately choose the ArcFace classification loss. We also find that the classic crop and resize data augmentation greatly hurts the performance, and accordingly devise data augmentation without scaling which benefits the performance. To study the scalability of the model, we scale this baseline to large models by several expanding techniques such as increasing input resolution and post-processing with re-ranking. Astonishingly, such techniques significantly boost the retrieval performance, which demonstrates that the proposed model has good scalability. In the future, we will investigate more robust losses for image-based patent retrieval.

\section*{Acknowledgement}
	This work was supported by the Start-up Research Fund of Southeast University under Grant RF1028623063. This work is also supported by the Big Data Computing Center of Southeast University.

\bibliographystyle{splncs04}
\bibliography{refs}
\end{document}